\documentclass[pdflatex,sn-mathphys-num]{sn-jnl}

\usepackage{graphicx}
\usepackage{multirow}
\usepackage{amsmath,amssymb,amsfonts}
\usepackage{amsthm}
\usepackage{mathrsfs}
\usepackage[title]{appendix}
\usepackage{xcolor}
\usepackage{textcomp}
\usepackage{manyfoot}
\usepackage{booktabs}
\usepackage{algorithm2e}
\usepackage{listings}
\usepackage{adjustbox}
\usepackage{resizegather}
\usepackage{subfigure}
\usepackage{multicol}
\usepackage{flushend}
\usepackage{makecell}
\usepackage{bm}

\usepackage[normalem]{ulem} 

\theoremstyle{thmstyleone}%
\theoremstyle{thmstyletwo}%

\theoremstyle{thmstylethree}%

\begin{document}

\title{Domain-Generalization to Improve Learning in Meta-Learning Algorithms}


\author*[1]{\fnm{Usman} \sur{Anjum}}\email{usman.anjum@ottawa.edu}

\author*[2]{\fnm{Chris} \sur{Stockman}}\email{stockmcr@mail.uc.edu}

\author[2]{\fnm{Cat} \sur{Luong}}\email{luongcn@mail.uc.edu}

\author[2]{\fnm{Justin} \sur{Zhan}}\email{zhanjt@ucmail.uc.edu}

\affil*[1]{\orgdiv{Department of Arts \& Sciences}, \orgname{Ottawa University}, \orgaddress{\city{Surprise}, \state{Arizona}, \postcode{85374}, \country{USA}}}
\affil*[2]{\orgdiv{Department of Computer Science}, \orgname{University of Cincinnati}, \orgaddress{\city{Cincinnati}, \state{Ohio}, \postcode{45221}, \country{USA}}}



\abstract{
This paper introduces Domain Generalization Sharpness-Aware Minimization Model-Agnostic Meta-Learning (DGS-MAML), a novel meta-learning algorithm designed to generalize across tasks with limited training data. DGS-MAML combines gradient matching with sharpness-aware minimization in a bi-level optimization framework to enhance model adaptability and robustness. We support our method with theoretical analysis using PAC-Bayes and convergence guarantees. Experimental results on benchmark datasets show that DGS-MAML outperforms existing approaches in terms of accuracy and generalization. The proposed method is particularly useful for scenarios requiring few-shot learning and quick adaptation, and the source code is publicly available at \href{https://github.com/AIResearchTopics/DGSharpMAML}{\textbf{GitHub}}.}


\maketitle

\section{Introduction}

 The pursuit of meta-learning ("learning to learn") involves training a model on a set of tasks and then using the knowledge acquired during training to generalize to new tasks that were not seen during the training phase. The goal of contemporary meta-learning is to enable models to adapt in a way similar to human learning, as it has been observed that children can often learn from just a single example. In contrast, traditional machine learning models typically required large amounts of data to achieve satisfactory performance. Current efforts in meta-learning focus on improving accuracy while reducing the amount of training data required. An effective meta-learning algorithm should be able to adapt to new and unseen tasks with limited samples and incorporate mechanisms to account for uncertainty, thereby preventing overfitting. This field is often associated with few-shot learning or domain generalization \cite{finn2017model}.

In recent years, meta-learning has gained significant attention, leading to the development of various algorithms. A landmark approach is Model-Agnostic Meta-Learning (MAML) \cite{finn2017model}. MAML uses stochastic gradient descent (SGD) \cite{bottou2018optimization} to solve its optimization problem and improve model generalization with limited training data. It is formulated as a bi-level optimization problem, where the outer-level learns a shared initialization across tasks, and the inner-level optimizes task-specific parameters. However, this bi-level structure introduces challenges: MAML is computationally expensive due to the need to compute second-order derivatives, and it inherits the limitations of SGD—such as the tendency to converge to sharp minima, which can hinder generalization.

To address these limitations, improved versions of MAML have been proposed. One such algorithm is SharpMAML \cite{abbas2022sharp}, which replaces SGD with Sharpness-Aware Minimization (SAM) \cite{foret2020sharpness} for optimization. SAM enhances generalization by minimizing the sharpness of the loss landscape. It achieves this by computing a perturbation in the model weights that maximizes the empirical risk, then minimizing the worst-case (perturbed) loss. In simpler terms, SAM minimizes the maximum loss in the vicinity of the current model parameters, rather than just the loss at a specific point. This approach has been shown to outperform SGD, and SharpMAML demonstrates higher accuracy than standard MAML. However, due to the complexity of the underlying min-max optimization, SAM only minimizes an approximation of the empirical risk through a perturbed loss. This approximation does not guarantee convergence, which has led to the development of further refinements to SAM.

One such refinement is Sharpness-Aware Gradient Matching (SAGM) \cite{wang2023sharpness}, a gradient-matching algorithm designed to improve generalization and optimization efficiency. SAGM leverages the concept of the surrogate gap \cite{zhuang2022surrogate}, aiming to perform no worse than SAM in terms of computational cost. When the surrogate gap approaches zero, it indicates that nearby model parameters yield similar loss values, suggesting a flatter loss surface. SAGM achieves generalization by jointly minimizing three objectives—empirical risk, perturbed loss, and surrogate loss—a strategy referred to as gradient matching. Although optimizing all three objectives can be challenging, \cite{wang2023sharpness} showed that this can be reformulated as minimizing the empirical and perturbed losses along with the angle between their gradients. This leads to convergence to flatter minima, which is desirable for generalization. Further details will be discussed in the Preliminaries and Methodology sections.

In this paper, we propose Domain-Generalization Sharpness-Aware Minimization MAML (DGS-MAML), which integrates the strengths of SAM and gradient matching. Our algorithm implements gradient matching within a bi-level optimization framework and updates parameters by implicitly aligning the empirical and perturbed losses. By minimizing both, we achieve low training loss; by reducing the surrogate gap, we narrow the performance gap between training and testing tasks. This enables us to converge to a flatter loss landscape and avoid sharp minima—without increasing computational cost.

In summary, our contributions are as follows.

\textbf{Formulation}: We formulate a novel algorithm, Domain Generalization Sharpness Aware Minimization Model Agnostic Meta-Learning, \textbf{DGS-MAML} that uses the concept of surrogate gap and gradient matching in conjunction with sharpness aware minimization to improve generalization, and do so using a bi-level algorithm structure. 

\textbf{Convergence Analysis}: We first perform theoretical analysis of the convergence of single-level SAGM and then extend the convergence analysis to bi-level optimization. We apply this convergence analysis to our algorithm DGS-MAML. We show that DGS-MAML does not fare worse than SAGM in convergence rate, and also show that it has improved convergence rate compared to both MAML and SharpMAML. We also provide PAC-Bayes analysis of DGS-MAML.

\textbf{Experimental Analysis}: We perform different experiments on benchmark datasets and show that DGS-MAML has better accuracy than the benchmark algorithms.


\section{Related Works}

\subsection{Model-agnostic meta learning}

MAML was first proposed in 2017 \cite{finn2017model} as a method of training any model utilizing gradient descent to update parameters, and so the method is applicable to a broad range of tasks, from classification to reinforcement learning. There have been many attempts to improve MAML, e.g. CAVIA \cite{zintgraf2019fast}, iMAML \cite{rajeswaran2019meta}, ANIL \cite{raghu2019rapid}, Matching Nets \cite{vinyals2016matching}, and REPTILE \cite{nichol2018first}. Kim et al., 2018 \cite{yoon2018bayesian} put forward BMAML, a Bayesian fast adaptation gradient-based meta-learning method. One advantage of this Bayesian approach is its avoidance of over-fitting at the meta level, although its efficiency left something to be desired. Closely related work to this \cite{finn2018probabilistic} extended the basic MAML algorithm to provide a distribution of models to sample for tasks unseen in training. Others \cite{grant2018recasting} recast MAML as an inference to distribution parameters in a hierarchical Bayesian context. Chen and Chen \cite{chen2022bayesian} compare Bayesian MAML with MAML and contend that Bayesian MAML can be theoretically shown to be superior, rather than only empirically. MAML's generalization has been studied in \cite{denevi2018learning} and \cite{farid2021generalization}. Other work \cite{wang2022st} leans into task ambiguity and uses it to learn solutions that allow a model to adapt to a new task. As mentioned above, an inspiration for our work here \cite{abbas2022sharp} leveraged sharpness aware minimization (SAM) in the MAML algorithm, leading to the bi-level SharpMAML, which will be discussed further in the Preliminaries. Another variation on MAML added momentum and adaptive learning to improve generalization \cite{anjum2025using}. Multiple recent studies have investigated MAML in the context of adversarial training (\cite{wang2021fast}, \cite{goldblum2020adversarially}, \cite{xu2021yet}), which, in brief, is a measure of how well a model will behave when an "adversary" perturbs the data in a way undetectable to humans. Fallah et al. (2021) \cite{fallah2021convergence} introduced stochastic gradient meta-reinforcement learning (SG-MRL), which is an extension of MAML to reinforcement learning problems. \cite{collins2020task} studied MAML in the "worst-case" scenario, when MAML was being applied to rare tasks, to gauge its task robustness. \cite{fallah2020personalized} studied how MAML performed in Federated Learning, which attempts to train a model that can serve as the initial model for multiple users that can then be adapted to each individual user to suit their data-driven needs. \cite{fallah2020convergence} studied the convergence of MAML and other gradient based methods, and also proposed a version of MAML that does not require a Hessian.

There have been multiple areas for implementation of meta-learning and bi-level algorithms. For example, one area of implementation has been in closed-loop supply chain (CLSC) management, especially in context of perishable goods \cite{zarreh2024integrating}. The Benders decomposition algorithm was developed as a type of bi-level problem as it used two efficient acceleration inequalities to tackle large-scale computational complexity \cite{makui2016accelerating, aazami2019benders}. A meta-heuristic algorithm based on cuckoo optimization algorithm (COA) was proposed to solve the problem in a short time and with a high quality \cite{goli2018accelerated}. The suppply chain problem was further studied as a four-level multi-objective algorithm in \cite{saeedi2017location}. Genetic algorithms and job shop scheduling algorithms have also been presented as a bi-level algorithm \cite{azad2019two, heydari2018minimizing}.

\subsection{Sharpness-Aware Minimization (SAM)}

Recent research has been done on the potential of SAM as an optimizer and its improved generalization performance. The SAM algorithm was proposed by Foret et. al. \cite{foret2020sharpness} in an effort to improve model generalizability by seeking a flat minimum. This proved to be a landmark on the topic, as since then, many variants of SAM have been proposed that not only improve on SAM but also improve the computation speed. Some methods that avoid SAM’s computational cost (one of few drawbacks to the method) come from adaptive learning, otherwise called "momentum" methods, some of which predate SAM. Some such methods are AdaGrad \cite{duchi2011adaptive}, Adam \cite{kingma2014adam}, and AMS-Grad \cite{reddi2019convergence} (which is a modification of Adam). Recently, another variant called AdaSAM [9] was proposed that combined both adaptive learning rate and momentum to improve the performance of SAM. Their results showed that using both the adaptive learning rate and momentum accelerates the convergence speed of SAM. Other algorithms such as mSAM \cite{behdin2023msam}, k-SAM \cite{ni2022k}, ESAM \cite{du2021efficient}, GSAM \cite{zhuang2022surrogate}, SAGM \cite{wang2023sharpness}, WASAM \cite{kaddour2022flat}, and ASAM \cite{kwon2021asam} have all modified the SAM algorithm to address deficiencies. Sharing a common goal with SAM is \cite{shu2021open}, which learns from different source domains to generalize well to unseen domains. 

mSAM and k-SAM improve on SAM by incurring a smaller calculation cost due to their reduced batch sizes of data. ESAM makes gradients sparse to reduce the computation cost in backpropagation. Like SAM, GSAM minimizes the loss function around a perturbed point and also makes use of a surrogate gap to further improve the minimization, although our parent paper \cite{wang2023sharpness} points out a drawback to GSAM. SAGM minimizes the empirical risk, the perturbed loss (i.e., the maximum loss within a neighborhood in the parameter space), and the gap between them to improve SAM's generalization capability and converge to the desired flat region. ASAM differs from SAM in that it scales the neighborhood adaptively such that the sharpness and parameter scaling are independent. WASAM brings a stochastic weight averaging metric to SAM to achieve its goal. AdaSAM proposed a novel momentum and adaptive learning method to improve on SAM's generalization and theoretically analyzed the convergence of an adaptive SAM learning algorithm.

\subsection{Gradient matching}

Our algorithm makes use of gradient matching, which—like MAML and SAM—has become popular. \cite{zhao2020dataset} employs gradient matching to condense large datasets into synthetic samples usable for training deep neural networks, promoting data-efficient learning. \cite{killamsetty2021grad} advanced GRAD-MATCH, targeting subsets whose gradients closely match those of the training or validation datasets. This approach reduces computational cost and demonstrates improved convergence bounds compared to competitors. \cite{shi2021gradient} proposed gradient matching between domains by maximizing the inner product between gradients from differing domains, using the algorithm Fish to approximate this optimization. \cite{jiang2023delving} provided a comprehensive overview of dataset condensation, investigating which gradients to match and using an adaptive learning approach to prevent overfitting. \cite{balles2022gradient} introduced a method to select a "coreset" of data that gradient matches those induced by training data for continual learning, aiming to retain knowledge from past updates. They employ a "rehearsal memory," where the model "remembers" certain data points when trained on new data. \cite{zeng2022gradient} studied gradient matching in federated learning to reduce domain discrepancies in classification tasks involving brain images.

\section{Preliminaries}

In this section, we review the basics of MAML and SharpMAML and describe some of the problems with optimization using MAML models. We also introduce the concept of domain generalization and gradient matching.

\subsection{Model Agnostic Meta-Learning (MAML)}

The goal of generalization is to train a model that can adapt to a new task using only a few data points. K-shot learning tasks are defined as $\{\mathcal{T}_m\}_{m=1}^K$ which is sampled from a distribution $p(\mathcal{T})$. Each task $m$ has a training set $\mathcal{D}_m^{train}=\{(x_i,y_i)\}_{i=1}^n$ and a validation set $\mathcal{D}_m^{val}=\{(x_i,y_i)\}_{i=1}^n$. The objective of MAML is to find the parameter $\theta$ (also called meta-model) for a model that maps $x$ to $y$ such that fine-tuning $\theta$ by small gradient updates can adapt the model to a new task. The loss function $\mathcal{L}$ provides task specific feedback. For a particular task $m$, the loss function is $\mathcal{L}(\theta;\mathcal{D}_m)$ where $\mathcal{D}_m$ is sampled from $\mathcal{D}$ as $\mathcal{D}_m^{train}$ and $\mathcal{D}_m^{val}$ for training and validation data sets respectively.

MAML is formulated as a bi-level optimization problem where at the lower-level task-specific parameters are optimized and at the higher level update $\theta$ is optimized. The MAML algorithm can be expressed as: 

\begin{equation} \label{eq:1}
    \min_{\theta} \sum_{\mathcal{T}_m \sim p(\mathcal{T})} \mathcal{L}_{\mathcal{T}_m} (f_{\theta_m'}) = 
    \sum_{\mathcal{T}_m \sim p(\mathcal{T})} \mathcal{L}_{\mathcal{T}_m} (f_{\theta - \beta\nabla_\theta \mathcal{L}_{\tau _m}(f_\theta)})
\end{equation}

It should be noted that in Equation \ref{eq:1}, $\theta_m' = \theta - \beta\nabla_\theta L_{\tau _m}(f_\theta)$ is the lower level update and the update of $\theta_m'$ is the higher level update. $f_{\theta_m'}$ and $f_{\theta_m}$ are models parameterized by $\theta_m'$ and $\theta_m$ respectively. $\mathcal{L}_{\mathcal{T}_m}$ is the loss for task $\mathcal{T}_m$ sampled from $p(\mathcal{T})$.

\subsection{Sharpness Aware Minimization (SAM) and SharpMAML}

SAM improves generalization by aiming for flatness in the loss landscape. SAM is based on the idea of flat-minima and approximates the minimal loss around the current parameter $\theta_t$. SAM first computes the worst-case perturbation $\epsilon$ that maximizes the loss within a given neighborhood $\alpha$, then minimizes the perturbed weight $\theta + \epsilon$. SAM can be expressed as the following minimax problem \cite{foret2020sharpness}:

\begin{equation} \label{eq:2}
    \min_\theta \max_{||\epsilon||_2 \leq \alpha} \mathcal{L}(\theta + \epsilon),
\end{equation}

in which $||\cdot||_2$ is the Euclidean norm. The sharpness of the loss function is defined as: 
\begin{equation}
    \max_{||\epsilon||_2 \leq \alpha} [\mathcal{L}(\theta + \epsilon; \mathcal{D}) - \mathcal{L}(\theta; \mathcal{D})]. \nonumber
\end{equation} 
SharpMAML is based on SAM and implemented in the service of bi-level model-agnostic meta-learning. In the inner loop (lower level), SharpMAML learns on training data sampled from $\mathcal{D}$, and in the outer loop (upper level), it is tested on validation data also from $\mathcal{D}$. SharpMAML can be expressed by the following equations:

\begin{equation} \label{eq:3}
   \theta_{t + 1} = \theta_t + \epsilon - \beta^l\nabla\mathcal{L}(\theta_t + \epsilon + \epsilon_m; \mathcal{D})), 
\end{equation}

\begin{equation} \label{eq:4}
    \epsilon_m = \alpha_l \frac{\nabla \mathcal{L}(\theta_t;\mathcal{D}_m^{train})}{||\nabla \mathcal{L}(\theta_t;\mathcal{D}_m^{train})||_2},
\end{equation}
and
\begin{equation} \label{eq:5}
    \epsilon = \alpha_u \frac{\nabla \sum_{m=1}^M \mathcal{L}(\theta_t + \epsilon_m;\mathcal{D}_m^{val})}{||\nabla \sum_{m=1}^M \mathcal{L}(\theta_t + \epsilon_m;\mathcal{D}_m^{val})||_2}.
\end{equation}
 
Equation \ref{eq:3} defines the lower level update step, with learning rate $\beta$ perhaps taking one value in the lower level and another in the upper level. Equation \ref{eq:4} defines the perturbation to be done to the parameter $\theta_t$ in the lower level. Equation \ref{eq:5} defines the upper level perturbation done to $\theta_t$, calculated using the results of the inner loop. The upper and lower level optimization is defined using the hyperparameters $\alpha_u$ and $\alpha_l$, radii of Euclidean balls.

\subsection{Domain Generalization and Gradient matching}

The goal of domain generalization is to enhance the generalization capability of a model from a seen domain to an unseen domain. Here we provide a brief primer on the concept of gradient matching and why one might use it. Gradient matching is based on the concept of vector dot products. A general formula for the dot product between two vectors is $a \cdot b = ||a||\cdot||b|| cos(\theta)$, where $\theta$ is the angle between the two vectors and $||\cdot||$ their magnitude, and $\cdot$ denotes ordinary multiplication with the magnitudes. When $\theta$ is near to $\frac{\pi}{2}$, that is, when $a$ and $b$ are nearly orthogonal, the cosine is nearly 0, and thus the inner product is very small. But when $\theta$ is near to 0, the cosine is near 1, and thus the inner product becomes $a \cdot b \approx ||a||\cdot||b||$. To continue with this toy example, in gradient matching algorithm, one vector, say $a$, is the gradient of the loss function, $\nabla \mathcal{L}(\theta; \mathcal{D})$, and the other, say $b$, is the gradient of the perturbed loss function, $\nabla \mathcal{L}_p(\theta; \mathcal{D})$.

The goal is to match the gradients of $\mathcal{L}$ at $\theta$ and at $\theta$ + some perturbation. This will maximize $a \cdot b$, as $||a||\cdot||b|| < ||a||^2$ when $||a|| > ||b||$. Thus, $\nabla \mathcal{L}(\theta; \mathcal{D}) \approx \nabla \mathcal{L}_p(\theta; \mathcal{D})$ entails that $\nabla \mathcal{L}(\theta; \mathcal{D}) \cdot \nabla \mathcal{L}_p(\theta; \mathcal{D})$ will be at its largest. We find that the objective of SAGM \cite{wang2023sharpness} is an effective way to achieve this goal.

\section{Methodology}

In this section we introduce our methodology that considers gradient matching to ensure the model converges to a flat loss region. We propose the DGS-MAML algorithm to achieve a good generalization performance.

\subsection{Domain Generalized SharpMAML (DGS-MAML) Algorithm}

\begin{figure}[h] 
  \centering
  \includegraphics[width=1.1\linewidth]{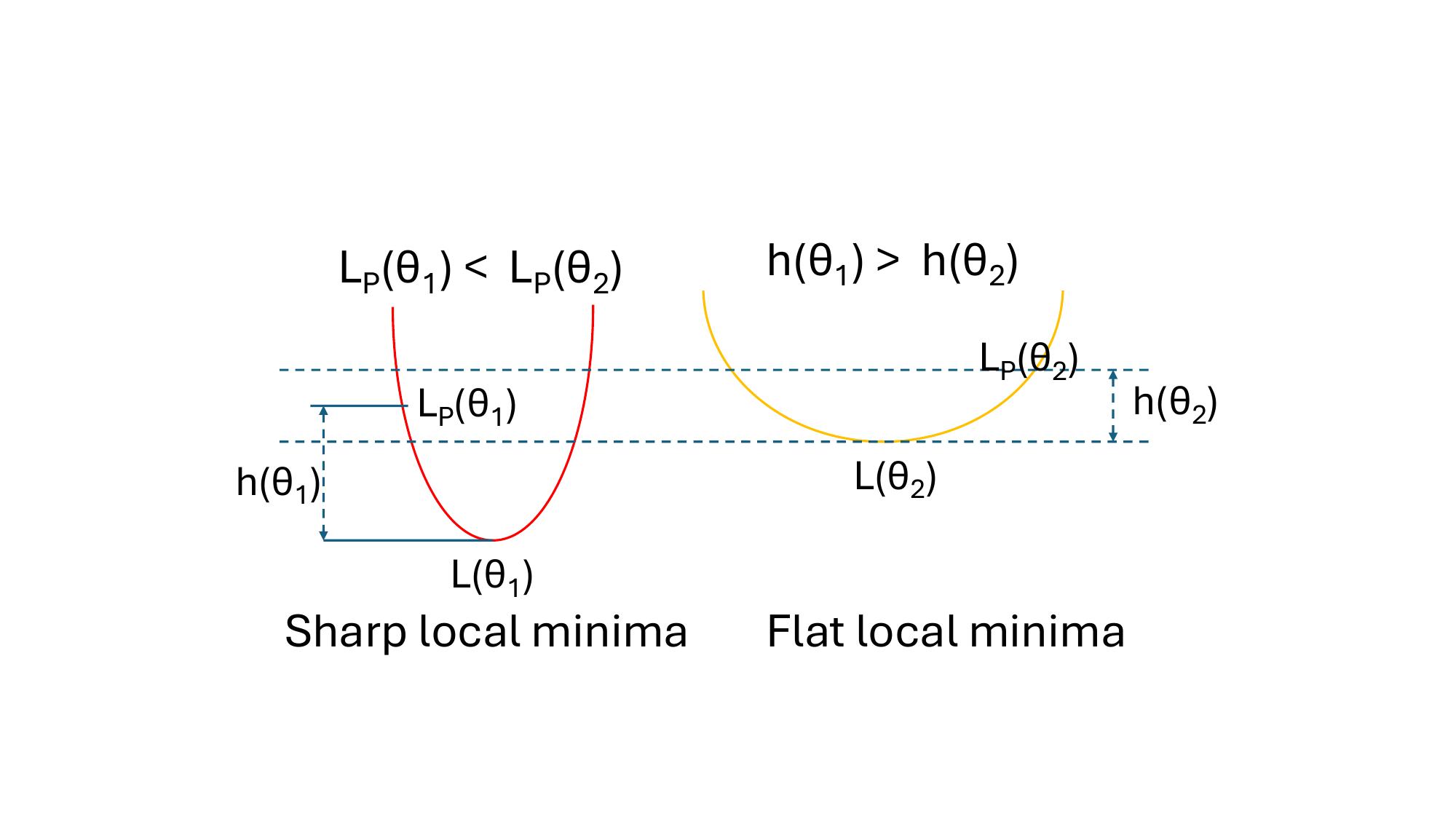}
  \caption{Comparison of sharp local minima $\theta_1$ (red) and flat minima $\theta_2$ (yellow): For the case $L_p(\theta_1;\mathcal{D}) < L_p(\theta_2;\mathcal{D})$ (where $L_p$ is the perturbed loss for the data $\mathcal{D}$), the SAM algorithm will return $\theta_1$ as the best solution, even though $\theta_2$ is the flatter local minimum. The surrogate gap $h(\theta)$ can better describe the sharpness of a loss surface and choose the correct local minima.}
    \label{fig:sagm}
\end{figure}

The aim of sharpness-aware learning is to find flat minima of a possibly non-convex loss landscape. This is done to increase model generalization. However, that solution may be very difficult to obtain, and there are several algorithms that have been devised to that end. Figure \ref{fig:sagm} shows the comparison between sharp local minima ($\theta_1$) and flat local minima ($\theta_2$). For the case $L_p(\theta_1;\mathcal{D}) < L_p(\theta_2;\mathcal{D})$ (where $L_p$ is the perturbed loss for the data $\mathcal{D}$), the SAM algorithm will return $\theta_1$ as the best solution, even though $\theta_2$ is the flatter local minimum. Hence, by introducing a surrogate gap $h(\theta)$, we can better describe the sharpness of a loss surface and choose the correct local minima.

Consequently, the recent work by \cite{wang2023sharpness} is based on this idea and showed that a smaller perturbed loss is not necessarily guaranteed to return a flat minimum. In such cases a sharp minimum may be found rather than a flat minimum.  Minimizing the perturbed loss was the goal of SAM, but it may pursue flatness to the point of returning a parameter that is more sensitive to data and may thus incur a higher loss. A similar approach to this, GSAM \cite{zhuang2022surrogate}, ran into the problem of pursuing gap minimization by raising the minimum rather than minimizing the perturbed loss; it thus sacrificed finding a lower minimum for finding a flatter one. Hence, researchers have proposed an alternate definition of the surrogate gap to better define sharpness (\cite{wang2023sharpness, zhuang2022surrogate}:
 \begin{equation} \label{eq:6}
     h(\theta) := \mathcal{L}_p(\theta; \mathcal{D}) - \mathcal{L}(\theta; \mathcal{D}),
 \end{equation}

 where $\mathcal{L}_p$ is the perturbed loss (very similar to SAM) and $\mathcal{L}$ is the standard training loss over training data $\mathcal{D}$ and parameter $\theta$. 
We define $\mathcal{L}_p(\theta_t ;\mathcal{D}) := \mathcal{L}(\theta_t + \epsilon - \delta \nabla\mathcal{L}(\theta_t;\mathcal{D});\mathcal{D})$. The surrogate gap measures the difference between the maximum loss within the neighborhood and the loss at the minimum point. Using the concepts of surrogate gap, \cite{wang2023sharpness} introduced the SAGM algorithm. We aim to take this idea from SAGM for a bi-level optimization meta-learning problem.

To achieve good generalization performance, SAGM meets two conditions: i) the loss within the minimum should be low; and ii) the minimum is within a flat surface. SAGM minimizes three things:

\begin{equation} \label{eq:6a}
    \min_{\theta} (\mathcal{L}(\theta;\mathcal{D}), \mathcal{L}_p(\theta, \mathcal{D}), h(\theta)).
\end{equation}

The above objective searches the low loss region that is within a flat surface. A low loss implies small training errors and the flat surface reduces the generalization gap between training and testing performance. 
To accomplish this objective, SAGM proposes a new objective: 
\begin{equation}\label{eq:7}
    \min_\theta \mathcal{L}(\theta; \mathcal{D}) + \mathcal{L}_p (\theta - \delta \nabla\mathcal{L}(\theta;\mathcal{D}); \mathcal{D}).
\end{equation}
Using this idea, we now have a minimax problem similar to that of SAM:

\begin{equation} \label{eq:8}
    \min_\theta \max_{||\epsilon||_2 \leq \alpha} \mathcal{L}(\theta_t + \epsilon - \delta \nabla\mathcal{L}(\theta_t;\mathcal{D});\mathcal{D}).
\end{equation}

Just as before from equations \ref{eq:3} and \ref{eq:4}, $\epsilon := \alpha\frac{\nabla\mathcal{L}(\theta_t,\mathcal{D})}{||\nabla\mathcal{L}(\theta_t,\mathcal{D})||}$, dropping the subscripts to $\alpha$ for convenience. Then, the equation above becomes:

\begin{equation} \label{eq:9}
    \min_\theta \max_{||\epsilon||_2 \leq \alpha} \mathcal{L}(\theta_t + (\frac{\alpha}{||\nabla\mathcal{L}(\theta_t,\mathcal{D})||} - \delta)\nabla\mathcal{L}(\theta_{t};\mathcal{D});\mathcal{D})
\end{equation}

From equation \ref{eq:9}, it can be observed that the computational cost of SAGM is the same as SharpMAML. It follows that our proposed DGS-MAML likewise fares no worse than the bi-level SharpMAML. This is because $\mathcal{L}(\theta; \mathcal{D})$ is already done during the normal calculations in SharpMAML, and everything else is merely arithmetic.

Equation \ref{eq:7} can be expanded around $\theta + \epsilon$ \cite{wang2023sharpness} to arrive at a rewriting of the original objective as:
\begin{align} \label{eq:10}
    \min_\theta \mathcal{L}(\theta; \mathcal{D}) + \mathcal{L}_p(\theta; \mathcal{D}) - \alpha\nabla\mathcal{L}_p(\theta; \mathcal{D}) \cdot \nabla\mathcal{L}(\theta; \mathcal{D})
\end{align}

\begin{figure}[h] 
  \centering
  \includegraphics[width=\linewidth]{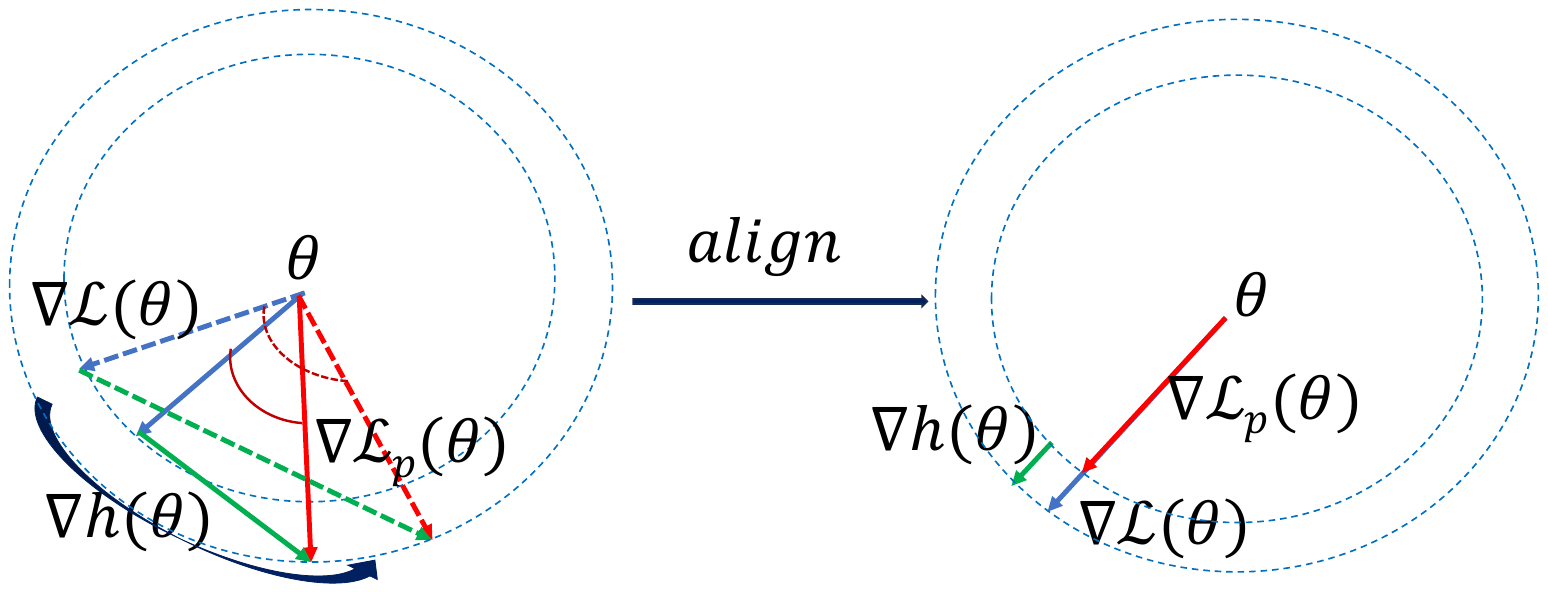}
  \caption{Components of $\nabla\mathcal{L}(\theta; \mathcal{D})$. When gradient directions of $\nabla\mathcal{L}(\theta)$ and $\nabla\mathcal{L}_p(\theta)$ align and the angle between them is 0, then $\mathcal{L}(\theta)$, $\mathcal{L}_p(\theta)$, and $h(\theta)$ can all be minimized efficiently.}
    \label{fig:gradient_match}
\end{figure}

It can be observed that when $\nabla \mathcal{L}_p(\theta; \mathcal{D})$ and $\nabla \mathcal{L}(\theta; \mathcal{D})$ are approximately equal, their inner product is greater than zero and achieves maximum value when the directions are equal. $\nabla h(\theta)$ in effect is a measure of how matched the gradients of the two losses are. Conflict in the gradients makes it difficult, if not impossible, to accomplish the objective in (\ref{eq:6}). When the gradients are equal, the loss function descends in the most efficient way and the dot product in (\ref{eq:10}) is maximized, which as a result minimizes the objective in (\ref{eq:6a}) and finds a flat region in the loss landscape. Matching the gradients of the $\mathcal{L}(\theta)$ and the $\mathcal{L}_p(\theta)$ avoids the problems mentioned above that SAM and GSAM are prone to running into. Minimizing $h(\theta)$ by matching the gradients takes more situations into account than SAM or GSAM did, since SAGM is not aiming solely at minimizing the perturbed loss, but is instead pursuing gradient matching, which as a byproduct will minimize the perturbed loss. To help illustrate and provide clarity, we provide Figure 2, inspired by Figure 2 of \cite{wang2023sharpness}. With the left side, $||\nabla \mathcal{L}(\theta)||$ (blue) and $||\nabla \mathcal{L}_p(\theta)||$ (red), radii of their respective circles, are in completely different directions. Focusing on only one of the components in the objective in (7) to minimize does not necessarily entail minimizing the others. For example, one could decrease $\nabla h(\theta)$ if they decreased $\nabla \mathcal{L}_p$ until it equaled $\nabla \mathcal{L}(\theta)$, but didn't actually touch $\nabla \mathcal{L}(\theta)$. This would illustrate pursuing flatness at the expense of accuracy. The other half of the figure depicts the case when the gradients are in the same direction. This is the only way to ensure that one efficiently minimizes all three terms in $(\mathcal{L}(\theta), \mathcal{L}_p(\theta), h(\theta))$. If the gradients were brought into conformity, to minimize $(\mathcal{L}(\theta)$ and $\mathcal{L}_p(\theta)$ would entail minimizing $h(\theta)$, and ensure that we have a flat loss landscape, which as a result will improve generalization.

Our parameter update steps are based on gradent matching but the structure is akin to SharpMAML's bi-level structure. In the inner loop we take a number of steps to set by the user and then in the outer loop, compute the new perturbation before using that new perturbation as part of the gradient matching update step. The algorithm for our paper is presented in Algorithm \ref{alg:DGS-MAML2}.

\RestyleAlgo{ruled}
\begin{algorithm} 
    \caption{DGS-MAML}
     \label{alg:DGS-MAML2}
    \KwData{$p(\mathcal{T}):$ distribution over tasks divided into training ($\mathcal{D}^{train}$) and validation ($\mathcal{D}^{val}$) sets}
    \KwResult{Updated Parameters}
    \textbf{Initialization:} meta-learning rate $\gamma$, perturbation radii $\alpha_l, \alpha_u > 0$, hyperparameter $\delta > 0$, starting parameter $\theta_0$ \\

    \For{t = 1,...T} {
        Sample tasks $\mathcal{T}_m \sim p(\mathcal{T})$ \\
        \For{$\mathcal{T}_m$} {
            Sample K samples from $\mathcal{D}_m^{train}$ \\
            Compute gradient $\nabla\mathcal{L}(\theta_{t - 1};\mathcal{D}_m^{train})$ \\
            Compute perturbation $\epsilon_m$ via eq (\ref{eq:4}) \\
            Compute new loss: $\mathcal{L}_{GM} = \mathcal{L}(\theta_{t - 1};\mathcal{D}_m^{train}) + \mathcal{L}(\theta_{t - 1} + \epsilon_m - \delta\nabla\mathcal{L}(\theta_{t - 1};\mathcal{D}_m^{train});\mathcal{D}_m^{train})$ \\
            Update weights: $\theta_{t} = \theta_{t - 1} - \gamma\nabla\mathcal{L}_{GM}$ \\ 
            Sample validation data $\mathcal{D}^{val}_m$\\
        }

        Compute $\sum_{m=1}^M\nabla\mathcal{L} (\theta_{t} + \epsilon_m;\mathcal{D}_m^{val})$ \\
        Compute perturbation $\epsilon$ via eq (\ref{eq:5}) \\
        Compute new loss: $\mathcal{L}_{GM} = \mathcal{L}(\theta_t;\mathcal{D}_m^{val}) + \mathcal{L}(\theta_t + \epsilon - \delta\nabla\mathcal{L}(\theta_t;\mathcal{D}_m^{val});\mathcal{D}_m^{val})$ \\
        Update weights: $\theta_{t} = \theta_{t - 1} - \gamma\nabla\mathcal{L}_{GM}$ \\ 
    }

\end{algorithm} 
 

\section{Convergence Analyses of SAGM and DGS-MAML}

In this section, we present an analysis of the update process used by our algorithm. As DGS-MAML is based on the SAGM algorithm, we first present an analysis of the convergence of SAGM \cite{wang2023sharpness}. It should be noted that the convergence of SAGM has not been done before. In a succeeding subsection we provide an analysis of DGS-MAML that incorporates variance in the computation of the gradient of the loss function over each sampled batch, with a third subsection containing the necessary lemmas that are appealed to in the main proofs. From our analysis, it will follow that the addition of an inner loop will not drag down the rate of convergence with respect to $T$. Here we merely state the result of the proof and it's corollary. 

\textbf{Theorem 1:
On assumptions 1 and 2 (see Section \ref{sec:5.1}), for fixed $\gamma$, and for $k = C\sqrt{d}(1 + \frac{\alpha}{C\sqrt{d}} - \delta)$, we arrive at the following convergence bound:}
\newline
\resizebox{\linewidth}{!}{
    \begin{minipage}{\linewidth}
\begin{align} \label{eq:11}
    & \frac{1}{T} \sum^{T-1}_{t=0} ||\nabla \mathcal{L}(\theta_{t})||^2 \leq \frac{\mathcal{L}(\theta_0) - \mathcal{L}(\theta)^{*}}{T} + & \frac{1}{T}(\gamma C\sqrt{d} k + (\gamma^2L + \gamma)C^2 d \nonumber \\
    & + \gamma^2 L k^2 - \frac{\gamma^2 L}{2} (k^2 + C^2 d - 2kC\sqrt{d})).
\end{align}
\end{minipage}
}

\textbf{Corollary 1: Convergence Rate of SAGM}

\textit{With $\alpha, \delta, \gamma$ all chosen to be $\mathcal{O}(\sqrt{\frac{1}{T}})$, we have that:}
\begin{equation}
    \frac{1}{T} \sum^{T-1}_{t=0} ||\nabla \mathcal{L}(\theta_{t})||^2 = \mathcal{O}({\frac{1}{T}}) \nonumber
\end{equation}

We observe also that we have improved on the convergence rate of SharpMAML \cite{abbas2022sharp}, which is one of $\mathcal{O}(\frac{1}{\sqrt{T}})$.

\subsection{Convergence Analysis of the SAGM Algorithm} \label{sec:5.1}

In this section, we prove the convergence rate of SAGM. The original work of \cite{wang2023sharpness} did not include a convergence analysis of the SAGM algorithm, and since DGS-MAML is a bi-level algorithm, we will begin the analysis of our algorithm with first considering the behavior of the single-level SAGM. Thus, what follows is original work on the convergence of SAGM. First, we state our assumptions.

\textbf{Assumption 1:} Lipschitz continuity. Let loss function $\mathcal{L}$ be such that $\nabla\mathcal{L}$ is $L$-Lipschitz continuous; that is, for parameters $x_1, x_2$, $||\nabla \mathcal{L}(x_1) - \nabla \mathcal{L}(x_2)|| \leq L||x_2 - x_1|| $.

\textbf{Assumption 2:}
Bounded stochastic gradient. We require that the stochastic gradient be uniformly bounded, i.e., $||\nabla \mathcal{L}(\theta_t; \mathcal{D}_m)||_{\infty} \leq C$, for $C \in \mathbb{N}$, and $\theta \in \mathbb{R}^d$.

We begin the proof of \cite{wang2023sharpness}'s algorithm with the descent inequality, which follows from our Lipschitz assumption:
\newline
\begin{align}
    \mathcal{L}(\theta_{t+1}) - \mathcal{L}(\theta_{t}) \leq \langle \nabla \mathcal{L}(\theta_{t}), \theta_{t + 1} - \theta_{t} \rangle + \frac {L}{2} ||\theta_{t + 1} - \theta_{t}||^2
\end{align}

Throughout the proof, we use $\nabla\mathcal{L}$ and $\nabla\mathcal{L}_p$ as shorthand for $\nabla \mathcal{L}(\theta_t; \mathcal{D}_m)$ and $\nabla \mathcal{L}_p(\theta_t +\hat{\epsilon} - \alpha \nabla \mathcal{L}(\theta_t; \mathcal{D}_m)); \mathcal{D}_m)$, respectively. Then, since $\theta_{t + 1} = \theta_{t} - \gamma \nabla \mathcal{L}_{SAGM}$ as per the SAGM update step:
\begin{align}
    & \mathcal{L}(\theta_{t + 1}) - \mathcal{L}(\theta_{t}) \leq \langle \nabla\mathcal{L}(\theta_{t}), - \gamma \nabla \mathcal{L}_{SAGM} \rangle \nonumber \\ 
    & + \frac{L}{2}||-\gamma \nabla \mathcal{L}_{SAGM}||^2 \\ 
    & \leq \langle \nabla \mathcal{L}, -\gamma \nabla \mathcal{L} \rangle + \langle \nabla \mathcal{L}, -\gamma \nabla \mathcal{L}_{p} \rangle + \frac{L\gamma^2}{2}||\nabla \mathcal{L}_{SAGM}||^2 \label{eq:15}\\ 
    & \leq \langle \nabla \mathcal{L}, - \gamma \nabla \mathcal{L}_{p} \rangle + \gamma ||\nabla\mathcal{L}||^2 + \frac{L\gamma^2}{2}||\nabla \mathcal{L} + \nabla \mathcal{L}p ||^2 \\ 
    & = \langle \nabla \mathcal{L},  - \gamma \nabla \mathcal{L}_{p} \rangle + \gamma ||\nabla \mathcal{L}||^2 + L\gamma^2(||\nabla \mathcal{L}||^2 \nonumber \\
    & + ||\nabla \mathcal{L}_{p}||^2 - \frac{1}{2}||\nabla \mathcal{L}_{p} - \nabla \mathcal{L}||^2) \\ 
    & = \langle \nabla \mathcal{L}, -\gamma \nabla \mathcal{L}_{p} \rangle + (\gamma^2L + \gamma)||\nabla \mathcal{L}||^2 \nonumber \\ 
    & + \gamma^2 L ||\nabla \mathcal{L}_{p}||^2 - \frac{\gamma^2L}{2}||\nabla h||^2 \\ 
    & \leq \nonumber \gamma ||\nabla\mathcal{L}||\cdot||\nabla \mathcal{L}_{p}|| + (\gamma^2 L + \gamma)||\nabla \mathcal{L}||^2 \nonumber \\ 
    & + \gamma^2 L ||\nabla \mathcal{L}_{p}||^2 - \frac{\gamma^2 L}{2} ||\nabla h||^2 
\end{align}
Then, by Lemma 3, we have:
\begin{align}
    & \mathcal{L}(\theta_{t + 1}) - \mathcal{L}(\theta_{t}) \leq \gamma C \sqrt{d}k + (\gamma^2 L + \gamma) C^2 d + \gamma ^2 L k^2 - \frac{\gamma^2 L}{2} ||\nabla h||^2 ,
\end{align}

where $k = C\sqrt{d}(1 + \frac{\alpha}{C\sqrt{d}} - \delta)$.
Then, by Lemma 4 (which bounds $||\nabla h||^2$):
\begin{align}
    & \mathcal{L}(\theta_{t + 1}) - \mathcal{L}(\theta_{t}) \leq \nonumber\\
    & \gamma C \sqrt{d}k + (\gamma^2 L + \gamma)C^2d + \gamma^2Lk^2 - \frac{\gamma^2 L}{2}(k^2 + C^2 d - 2kC\sqrt{d}), \label{eq:21}
\end{align}

where $k = C\sqrt{d}(1 + \frac{\alpha}{C\sqrt{d}} - \delta)$.
Then, by telescoping and taking the average over $T$ steps, we arrive at the conclusion:
\newline
\begin{align}
    & \frac{1}{T} \sum^{T-1}_{t=0} ||\nabla \mathcal{L}(\theta_{t})||^2 \leq 
    \frac{\mathcal{L}(\theta_0) - \mathcal{L}(\theta)^*}{T} + 
    \frac{1}{T}(\gamma C\sqrt{d} k + (\gamma^2L + \gamma)C^2 d \nonumber \\
    & + \gamma^2 L k^2 - \frac{\gamma^2 L}{2} (k^2 + C^2 d - 2kC\sqrt{d})),
\end{align}

where $\mathcal{L}(\theta)^* := \mathcal{L}(\theta_T)$, the lowest value taken by the loss function. This completes the proof. 

\subsection{Convergence Analysis of DGS-MAML} \label{sec:5.2}

In this section, we present the convergence analysis of DGS-MAML. In DGS-MAML, the introduction of an inner loop slightly changes the derivation process for the convergence analysis. However, we show that the convergence rate remains the same. We require an additional assumption of bounded variances.

\textbf{Assumption 3:}
Take $\widetilde{\mathcal{L}}(\theta_t)$ as shorthand for $\mathcal{L}(\theta_t ; \xi_m)
$, where $\xi_m \sim D_{m}$ is sampled data from the dataset for the task at hand. Assume
$\widetilde{\nabla}\mathcal{L}(\theta_{t}; \xi_m), \widetilde{\nabla}\mathcal{L}_{p}(\theta_{t}; \xi_m)$ are unbiased estimators of 
$\nabla \mathcal{L}(\theta_t)$, $\nabla \mathcal{L}_p(\theta_{t}; \mathcal{D}_m)$, i.e.:
\begin{align}
    \mathbb{E}[\widetilde{\nabla}\mathcal{L}(\theta_{t})] = \nabla \mathcal{L} (\theta_{t};\mathcal{D}_m), \nonumber
\end{align}
and
\begin{align}
    \mathbb{E}[||\widetilde{\nabla}\mathcal{L}(\theta_{t}) - \nabla \mathcal{L}(\theta_{t}; \mathcal{D}_m)||^2] \leq \sigma_{1}^2, \nonumber
\end{align}
and 
\begin{align}
    \mathbb{E}[\widetilde{\nabla}\mathcal{L}_{p}(\theta_{t})] = \nabla \mathcal{L}_p(\theta_{t}; \mathcal{D}_m), \nonumber
\end{align}
and 
\begin{align}
    \mathbb{E}[||\widetilde{\nabla}\mathcal{L}_p(\theta_{t}) - \nabla \mathcal{L}_{p}(\theta_{t}; \mathcal{D}_m)||^2] \leq \sigma^{2}_{2} .\nonumber  
\end{align}

We provide this proof incorporating variances to our prior assumptions. Due to the bi-level nature of DGS-MAML, pure SAGM assumptions do not always hold, or at least are insufficient to serve as an exhaustive base of assumptions. This is due to the batches of data containing outliers when compared to the distribution of the task dataset as a whole. We begin the DGS-MAML proof starting from a slightly changed (14): 

\begin{align}
    & \widetilde{\mathcal{L}}(\theta_{t + 1}) \leq \widetilde{\mathcal{L}}(\theta_{t}) + \gamma \langle \widetilde{\nabla} \mathcal{L}(\theta_{t}), \widetilde{\nabla} \mathcal{L}_{SAGM} \rangle + \frac{L}{2} || -\gamma \widetilde{\nabla} \mathcal{L}_{SAGM}||^2 
\end{align}

Similar to (\ref{eq:15}), we have:
\begin{align}
    & \mathbb{E}[\widetilde{\mathcal{L}}(\theta_{t + 1} \, | \, \mathcal{F}_{t}] \leq \widetilde{\mathcal{L}}(\theta_{t}) + \gamma \mathbb{E}[\langle \widetilde{\nabla} \mathcal{L}, \widetilde{\nabla}\mathcal{L}_{SAGM} \rangle \, | \, \mathcal{F}_{t}] \nonumber \\ 
    & + \frac{L \gamma^2}{2} \mathbb{E}[||\widetilde{\nabla}\mathcal{L}_{SAGM}||^2 \, | \, \mathcal{F}_{t}]
\end{align}

Then by Lemma 1:
\begin{align}
    & \mathbb{E}[\widetilde{\mathcal{L}}(\theta_{t + 1}) \, | \, \mathcal{F}_{t}] \leq \widetilde{\mathcal{L}}(\theta_{t}) + \gamma (C^2d + \sigma^{2}_{1} \nonumber \\
    & + \sqrt{C^2 d + \sigma_{1}^2} \sqrt{k^2 + \sigma_{2}^2}) + \frac{L \gamma^2}{2} \mathbb{E}[|| \widetilde{\nabla} \mathcal{L}_{SAGM}||^2 \, | \, \mathcal{F}_{t}].
\end{align}

And, by Lemma 2: 
\begin{align}
    & \mathbb{E}[\widetilde{\mathcal{L}}(\theta_{t + 1}) \, | \, \mathcal{F}_{t}] \leq \widetilde{L}(\theta_{t}) + \gamma (C^2 d + \sigma_{1}^2 + \sqrt{(C^2 d + \sigma_{1}^2)(k^2 + \sigma^{2}_{2})}) \nonumber \\
    & + L \gamma^2 ( C^2 d + \sigma_{1}^2 + k^2 + \sigma_{2}^2 + 2\sqrt{(C^2 d + \sigma_{1}^2)(k^2 + \sigma_{2}^2)} ).
\end{align}

Then, taking the expectation over all t: 
\begin{align}
    & \mathbb{E}[\widetilde{L}(\theta_{t + 1}) - \widetilde{L}(\theta_{t})] \leq \gamma (C^2 d + \sigma_{1}^2 + \sqrt{(C^2 d + \sigma_{1}^{2})(k^2 + \sigma_{2}^{2})} ) \nonumber \\
    & + L \gamma^2 (C^2 d + \sigma_{1}^2 + k^2 + \sigma_{2}^{2} + 2\sqrt{(C^2 d + \sigma_{1}^{2})(k^2 + \sigma_{2}^{2})})
\end{align}

And, by telescoping and averaging over T iterations we arrive at: 
\begin{align}
    & \frac{1}{T} \sum^{T - 1}_{t = 0} \mathbb{E}||\widetilde{\nabla} \mathcal{L}(\theta_{t}; \xi_{m})||^2 \leq \frac{\mathcal{L}(\theta_0) - \mathcal{L}(\theta)^{*}}{T} + \frac{1}{T}( \gamma (C^2 d + \sigma_{1}^2 \nonumber \\
    & + \sqrt{(C^2 d + \sigma_{1}^{2})(k^2 + \sigma_{2}^{2})} ) \nonumber \\ 
    & + L \gamma^2 (C^2 d + \sigma_{1}^2) + (k^2 + \sigma_{2}^{2} + 2\sqrt{(C^2 d + \sigma_{1}^{2})(k^2 + \sigma_{2}^{2})})),
\end{align}
where $\mathcal{L}(\theta)^{*} := \mathcal{L}_T$, the smallest loss value. 
\newline
This completes the proof, and we have a second theorem and associated corollary:

\textbf{Theorem 2:
On our three assumptions, for fixed $\gamma$, and for $k = C\sqrt{d}(1 + \frac{\alpha}{C\sqrt{d}} - \delta)$, we arrive at the following convergence bound:}
\newline
\resizebox{\linewidth}{!}{
    \begin{minipage}{\linewidth}
\begin{align} \label{eq: Theorem2}
    & \frac{1}{T} \sum^{T - 1}_{t = 0} \mathbb{E}||\widetilde{\nabla} \mathcal{L}(\theta_{t}; \xi_{m})||^2 \leq \frac{\mathcal{L}(\theta_0) - \mathcal{L}^{*}}{T} + \frac{1}{T}( \gamma (C^2 d + \sigma_{1}^2 \nonumber \\
    & + \sqrt{(C^2 d + \sigma_{1}^{2})(k^2 + \sigma_{2}^{2})} ) \nonumber \\ 
    & + L \gamma^2 (C^2 d + \sigma_{1}^2) + (k^2 + \sigma_{2}^{2} + 2\sqrt{(C^2 d + \sigma_{1}^{2})(k^2 + \sigma_{2}^{2})})).
\end{align}
\end{minipage}
}
Our second corollary is:

\textbf{Corollary 2: Convergence Rate of DGS-MAML}\newline
\textit{With $\alpha, \delta, \gamma$ all chosen to be $\mathcal{O}(\sqrt{\frac{1}{T}})$, we have that: \newline
$\frac{1}{T} \sum^{T-1}_{t=0} \mathbb{E}||\widetilde{\nabla} \mathcal{L}(\theta_{t}; \xi_{m})||^2  = \mathcal{O}({\frac{1}{T}})$.
} 

\section{DGS-MAML PAC-Bayes Analysis}
In this section we provide a PAC-Bayes generalization proof for our algorithm. To start, we state the definition of the PAC-Bayes theorem, taken from (\cite{farid2021generalization}, \cite{abbas2022sharp}). We will start with a definition of uniform stability.

Suppose $D = \{s_1, s_2, ..., s_n\} \in S^m$, where $S^m$ is the sample space for the $m$th task, is one dataset of $n$ elements of $S$. Suppose also that $D' = \{s_1, ..., s_{j}',..., s_n\}$ is identical to $D$ in all elements except for the $j$th element $s_j$ in $D$ is some $s_{j}' \in S$ instead. 

\textbf{Definition 1} (Uniform Stability \cite{farid2021generalization}) \textit{A deterministic algorithm A is called $U$-uniformly stable, for $U > 0$, with respect to loss $\mathcal{L}$ if $\forall s \in S, \forall D \in S^m, \forall j \in \{1, ..., n\}$, and all distributions $P_{\theta}$ over initializations, $U$ is the smallest number such that the following holds:}
\begin{equation} \label{Uniform stability def}
    \mathbb{E}_{\theta \sim P_\theta}|\mathcal{L}(h_{A(\theta; \mathcal{D})}; s) - \mathcal{L}(h_{A(\theta; \mathcal{D}')}; s)| \leq U
\end{equation}

\textbf{Theorem 3} (PAC-Bayes Uniform Stability \cite{abbas2022sharp}) \textit{On the assumption that the loss is bounded $0 \leq \mathcal{L}(h_{A(\theta, \mathcal{D})}, \mathcal{D}) \leq 1$ for arbitrary $h$ in the hypothesis space and any $\mathcal{D} \in S$, where $h$ is learned via a $U$-uniformly stable algorithm $A$ with arbitrary data-independent prior distribution $P_{\theta}$ over initializations for $\theta$, and $\psi \in (0, 1)$, the following holds with probability $1 - \psi$ over a sampling of the meta-training dataset $D \sim S$ and arbitrary posterior distribution $Q_{\theta}$:}
\begin{equation} \label{PAC theorem}
   \begin{aligned}[b]
    & \mathbb{E}_{\mathcal{D} \sim S} \mathbb{E}_{\theta \sim Q_{\theta}} \mathcal{L}(h_{A(\theta; \mathcal{D})}, \mathcal{D}) \\
    & \leq \frac{1}{K} \sum_{i = 1}^{K} \mathbb{E}_{\theta \sim Q_{\theta}} \mathcal{L}(h_{A(\theta; \mathcal{D}_i)}, \mathcal{D}_i) + \sqrt{\frac{D_{KL}(Q_{\theta}||P_\theta) + ln(\frac{2\sqrt{K}}{\psi})}{2K}} + U 
    \end{aligned}
\end{equation}
\textit{Note that $|\mathcal{D}| = K$ here.}

In our case, we have that some parameter $\hat{\theta} \in \mathbb{R}^d$ is the learned hypothesis and is a local minimizer. The following assumption needed for our proof will follow: 
\begin{equation}
    \mathbb{E}_{\mathcal{D} \sim S}\mathcal{L}(\hat{\theta}; \mathcal{D}) \leq \mathbb{E}_{\epsilon \sim \mathcal{N}(\textbf{0}, (\alpha^2 + \delta^2)\textbf{I})}\mathbb{E}_{\mathcal{D} \sim S} \mathcal{L}(\hat{\theta} + \epsilon; \mathcal{D}).
\end{equation}

In proving our PAC-Bayes bound, the key term to bound is the Kullback-Liebler divergence. Since it is assumed that the prior and posterior distributions are arbitrary, we choose prior $P_{\theta} \sim \mathcal{N}(\textbf{0}, \sigma_p^2\textbf{I})$ and posterior $Q_{\theta} \sim \mathcal{N}(\hat{\theta}, (\alpha^2 + \delta^2)\textbf{I})$. In the case of two Normal distributions, the KL divergence is calculated as 
\begin{equation} 
    \begin{aligned}
     & D_{KL}(Q||P_\theta) = \frac{1}{2}[\text{tr}(\mathbf{\Sigma}_P^{-1}\mathbf{\Sigma_Q}) + (\boldsymbol{\mu}_P - \boldsymbol{\mu}_Q)^{\text{T}}\mathbf{\Sigma_P^{-1}}(\boldsymbol{\mu}_P - \boldsymbol{\mu}_Q) - d + \text{ln}(\frac{|\mathbf{\Sigma}_P|}{|\mathbf{\Sigma}_Q|})]. 
    \end{aligned}
\end{equation}
By plugging in the assumed information, and carrying out the operations of trace and determinant, we have:
\begin{align}
     & D_{KL}(Q_{\theta}||P_{\theta}) = \frac{1}{2}[\text{tr}(\frac{\alpha^2 + \delta^2}{\sigma_p^2}\textbf{I}) + \hat{\theta}^T(\frac{1}{\sigma_p^2}\textbf{I})\hat{\theta} - d + \text{ln}(\frac{|\sigma_p^2 \textbf{I}|}{|(\alpha^2 + \delta^2)\textbf{I}|} ] \\
     & = \frac{1}{2}[\frac{(\alpha^2 + \delta^2)d + ||\hat{\theta}||^2}{\sigma_p^2} - d + d\text{ln}(\frac{\sigma_p^2}{\alpha^2 + \delta^2}) ]
\end{align}
It is required that $D_{KL} > 0$, and to ensure this, we require that each term be positive. Therefore in the $d(\text{ln}(\frac{\sigma_p^2}{\alpha^2 + \delta^2}) - 1)$ term, it will be required that the ln be greater than or equal to 1. This occurs when its argument is greater than its base, and so we have a lower bound of $\sigma_p^2$:

\begin{equation} \label{eq:36}
    \frac{\sigma_p^2}{\alpha^2 + \delta^2} \geq e \nonumber \\
    \sigma_p^2 \geq  (\alpha^2 + \delta^2)e
\end{equation}

In \cite{abbas2022sharp}, it was shown that setting $\alpha$ close to $0$ reduces SharpMAML to the original MAML, which leads to looser generalization bounds. Since the generalization bound in SharpMAML depends on $\alpha$, a larger $\alpha$ results in a tighter (i.e., smaller) generalization error. In DGS-MAML, we introduce an additional $\delta$ term in the bound (Equation \ref{eq:36}). This added term further reduces the generalization error, thereby improving the upper bound on generalization and enhancing the performance beyond that of SharpMAML.

\section{Experiment}

In this section we perform different experiments on various datasets and compare with other state-of-the-art meta-learning algorithms like MAML and SharpMAML.

\subsection{Experimental Setup}
The neural network backbone architecture is based on the same architecture in \cite{vinyals2016matching}. In this architecure there are 
4 modules with $3 \times 3$ convolution layers with 64 filters, a batch normalization layer, a ReLU, and a $2 \times 2$ max-pooling layer. We use the Adam optimizer with learning rates of [0.1, 0.01, 0.001] for DGS-MAML and SAM. Both $\alpha_l$ and $\alpha_u$ values were equal. 

We used four datasets that have been used previously for image recognition tasks. The dataset used were Mini-Imagenet \cite{ravi2016optimization}, DoubleMNIST \cite{mulitdigitmnist}, and TripleMNIST \cite{mulitdigitmnist} and Omniglot \cite{lake2011one}. The Omniglot dataset consists of 20 instances of 1623 characters from 50 different alphabets. The Mini-Imagenet dataset has a train/ validation/ test split of 64/12/24. The DoubleMNIST and TripleMNIST have a train/ validation/ test split of 640/200/160 and  64/16/20 respectively. The Omniglot dataset is split according to \cite{vinyals2016matching}.

We used N-way K-shot learning to train and test the model, i.e. we provided the model with K-samples of N-classes. The dataset processing was done using the Torchmeta library \cite{deleu2019torchmeta}.

We compared our method to the following algorithms: MAML \cite{finn2017model}, SharpMAML(both) \cite{abbas2022sharp}, CAVIA \cite{zintgraf2019fast} \footnotemark[1], REPTILE \cite{nichol2018first} \footnotemark[1], Matching Networks \cite{vinyals2016matching} \footnotemark[2], and Protonet \cite{snell2017prototypical} \footnotemark[2].  

\footnotetext[1]{Implemented using open source code from \url{https://github.com/sungyubkim/GBML/tree/master}}
\footnotetext[2]{Implemented using open source code from \url{https://github.com/tristandeleu/pytorch-meta/tree/master/examples}}

\begin{table*}[htp]
\caption{Experiment Results - Using 5-way 1-shot and 5-way 5-shot Mini-Imagenet data. The best values for accuracy are in bold.}
\begin{center}
\begin{tabular}{c|c|c}
\hline
\textbf{Method} & \textbf{5-way 1-shot} & \textbf{5-way 5-shot}\\
\hline
Matching Networks & $0.4033$ & $0.5067$\\
\hline
CAVIA & $0.4599$ & $0.5661$\\
\hline
Reptile & $0.4181$ & $0.5278$\\
\hline
Protonet & $0.4390$ & $0.5884$\\
\hline
MAML & $0.4463$ & $0.5729$\\
\hline
\textbf{$SharpMAML_{both} (\alpha = 0.05)$} & $0.4509$ & $0.5759$\\
\hline
\textbf{$SharpMAML_{both} (\alpha = 0.005)$} & $0.4351$ & $0.5677$\\
\hline
\textbf{$DGS-MAML (\alpha = 0.05)$} & $\textbf{0.4665} (\delta = 2.0)$ & $\textbf{0.6360} (\delta = 0.01) $\\
\hline
\textbf{$DGS-MAML (\alpha = 0.005)$} & $0.4607 (\delta = 0.005)$ & $0.6246 (\delta = 0.01)$\\
\hline
\end{tabular}
\label{tab1:mini}
\end{center}
\end{table*}

\begin{table}[htp]
\caption{Experiment Results - Using 10-way 1-shot and 10-way 5-shot Mini-Imagenet data. The best values for accuracy are in bold.}
\centering
\begin{tabular}{c|c|c}
\hline
\textbf{Method} & \textbf{10-way 1-shot} & \textbf{10-way 5-shot}\\
\hline
MAML & $0.2160$ & $0.3621$\\
\hline
\textbf{$SharpMAML_{both} (\alpha = 0.05)$} & $0.2703$ & $0.3899$\\
\hline
\textbf{$SharpMAML_{both} (\alpha = 0.005)$} & $0.2740$ & $0.3846$\\
\hline
\textbf{$DGS-MAML (\alpha = 0.05, \delta = 0.1)$} & $\textbf{0.2882}$ & $0.4213$\\
\hline
\textbf{$DGS-MAML (\alpha = 0.005)$} & $0.2879 (\delta = 0.006)$ & $\textbf{0.4264} (\delta = .01)$\\
\hline
\end{tabular}
\label{tab2:mini}
\end{table}

\begin{table*}[htp]
\caption{Experiment Results - Using 20-way 1-shot and 20-way 5-shot Omniglot data. The best values for accuracy are in bold.}
\begin{center}
\begin{tabular}{c|c|c}
\hline
\textbf{Method}& \textbf{20-way 1-shot} & \textbf{20-way 5-shot}\\
\hline
Matching Networks & $0.6760$ & $0.6329$\\
\hline
MAML & $0.8884$ & $0.9571$\\
\hline
\textbf{$SharpMAML_{both} (\alpha = 0.005)$} & $0.8948$ & $0.9601$\\
\hline
\textbf{$SharpMAML_{both} (\alpha = 0.05)$} & $0.8945$ & $0.9581$\\
\hline
\textbf{$DGS-MAML (\alpha = 0.005, \delta = 0.1)$} & $\textbf{0.8982}$ & $0.9602$\\
\hline
\textbf{$DGS-MAML (\alpha = 0.05)$} & $0.8952  (\delta = 0.5)$ & $\textbf{0.9604} (\delta = 0.1)$\\
\hline
\end{tabular}
\label{tab3:omni}
\end{center}
\end{table*}

\begin{table}[htp]
\caption{Experiment Results - Using 10-way 1-shot and 10-way 5-shot TripleMNIST data. The best values for accuracy are in bold.}
\centering
\begin{tabular}{c|c|c}
\hline
\textbf{Method}& \textbf{20-way 1-shot} & \textbf{20-way 5-shot}\\
\hline
MAML & $0.9017$ & $0.9851$\\
\hline
\textbf{$SharpMAML_{both} (\alpha=0.005)$} & $0.9014$ & $0.9827$\\
\hline
\textbf{$SharpMAML_{both} (\alpha=0.05)$} & $0.9136$ & $0.9844$\\
\hline
\textbf{$DGS-MAML (\alpha=0.005)$} & $0.9131 (\delta=0.1)$ & $0.9842 (\delta=0.05)$\\
\hline
\textbf{$DGS-MAML (\alpha=0.05)$} & $\textbf{0.9171} (\delta=0.1)$ & $\textbf{0.9852} (\delta=0.05)$\\
\hline
\end{tabular}
\label{tab4:triple}
\end{table}

\begin{table}[htp]
\caption{Experiment Results - Using 20-way 1-shot and 20-way 5-shot DoubleMNIST data. The best values for accuracy are in bold.}
\centering
\begin{tabular}{c|c|c}
\hline
\textbf{Method}& \textbf{10-way 1-shot} & \textbf{10-way 5-shot}\\
\hline
MAML & $0.9631$ & $0.9898$\\
\hline
\textbf{$SharpMAML_{both} (\alpha = 0.005)$} & $0.9639$ & $\textbf{0.9931}$\\
\hline
\textbf{$SharpMAML_{both} (\alpha = 0.05)$} & $0.9641$ & $0.9921$\\
\hline
\textbf{$DGS-MAML (\alpha = 0.005)$} & $0.9641 (\delta=0.1)$ & $0.9910 (\delta=0.1)$\\
\hline
\textbf{$DGS-MAML (\alpha=0.05)$} & $\textbf{0.9661} (\delta=0.1)$ & $0.9930 (\delta=0.05)$\\
\hline
\end{tabular}
\label{tab5:double}
\end{table}

\subsection{Experimental Results}
We present our results in Tables \ref{tab1:mini}, \ref{tab2:mini}, \ref{tab3:omni}, \ref{tab4:triple} and \ref{tab5:double}. The corresponding $\delta$ value that gave the best results is also reported in the table. The results reveal using domain generalization, we improve the accuracy of sharpMAML and we are able to outperform the state-of-the-art algorithms in all the datasets.  

The results show that by adding domain generalization and fine-tuning the $\delta$ parameter, DGS-MAML is able to improve the generalization performance of SharpMAML and performs better than the state-of-the-art methods. For tasks that already have high accuracy, the increase is not so significant, but for tasks with relatively low accuracy (like Mini-Imagenet) the increase in accuracy is quite significant.

\begin{table}[htp]
\caption{Training Run-time Comparison/epoch - batch size=100}
\centering
\begin{tabular}{c|c|c|c}
\hline
 $Runtime \, (secs)$ & $MAML$ & $SharpMAML$ & $DGS-MAML$\\
\hline
\textbf{Mini-Imagenet} & $70-75$ & $60-65$ & $60-70$\\
\hline
\hline
\textbf{Omniglot} & $60-70$ & $60-70 $ & $60-70$\\
\hline
\end{tabular}
\label{tab7}
\end{table}

\begin{table*}[htp]
\caption{Experiment Results - hyperparameter tuning}
\centering
\resizebox{\linewidth}{!}{
\begin{tabular}{c|c|c|c|c|c|c|c|c}
\hline
\textbf{Data} & $\delta = 0.0005$ & $\delta = 0.001$ & $\delta = 0.009$ & $\delta=0.01$ & $\delta=0.02$ & $\delta=0.03$ & $\delta=0.1$ & $\delta = 2.0$\\
\hline
\textbf{Mini-Imagenet} \textbf{(5x1)} & $0.4476$ & $0.4545$ & $0.4538$ & $0.4507$ & $0.4517$ & $0.4482$ & $0.4521$ & $0.4651$\\
\hline
\end{tabular}
}
\label{tab5}
\end{table*}

\begin{table*}[htp]
\caption{Experiment Results - hyperparameter tuning}
\begin{center}
\begin{tabular}{c|c|c|c|c|c}
\hline
\textbf{Data} & $\delta = 0.002$ & $\delta = 0.004$ & $\delta = 0.006$ & $\delta = 0.008$ & $\delta = 0.01$ \\
\hline
\textbf{Mini-Imagenet} \textbf{(5x5)} & $0.6219$ & $0.6146$ & $0.6249$ & $0.6134$ & $0.6360$ \\
\hline
\end{tabular}
\label{tab6}
\end{center}
\end{table*}

\subsection{Ablation Studies}

Using Mini-Imagenet with 5-way 1-shot and 5-way 5-shot configurations, we also ran experiments on the changing $\delta$ value. The results are shown in Table \ref{tab5} and Table \ref{tab6}. In Table \ref{tab5} the results show that we get the best accuracy at a specific $\delta$ value and accuracy is lower at all other values. In Table \ref{tab6}, for fixed $\alpha$ we observe accuracy when DGS-MAML is run on a range of $\delta$ values, some of which seem not to make significant impact, although there is a clear winner. These results shows that fine-tuning $\delta$ is crucial in getting the best accuracy, and any future work using DGS-MAML should aim to pursue a method that automatically finds the best $\delta$ value. Suggestions for future work would include hyperparameter optimization for $\alpha$ and $\delta$. It should we noted that we did not change the value of $\alpha$ because it has been studied in previous literature and our algorithm mainly uses the $\delta$ parameter. However, our results do show that in general DGS-MAML performs best at $\alpha=0.05$.

We also ran experiments to compare the run-time per epoch for MAML, SharpMAML and DGS-MAML. The results are reported in Table \ref{tab7}. The increase in runtime for DGS-MAML is very low compared to SharpMAML.

\section{Discussion}

This section explores the practical utility of DGS-MAML, emphasizing how its ability to rapidly adapt and generalize makes it particularly effective for real-world challenges.

DGS-MAML improves model generalization, making it more efficient in scenarios where data is scarce, and models need to quickly adapt to new tasks without having the need to retrain the entire model. DGS-MAML can be used in natural language processing for event detection when limited text data is available \cite{DBLP:conf/iscram/LazregAZG20, DBLP:conf/iscram/AnjumZK21, DBLP:journals/osnm/AnjumZK22, nkhata2023sentiment, anjum2025weighted, anjum2023a}. Similarly, in cyberbullying detection, DGS-MAML could enable few-shot learning to identify harmful content \cite{DBLP:conf/bigdataconf/GuoAZ22, nkhata2023sentiment} instead of relying on data augmentation techniques.

In autonomous systems (like cars and robots), DGS-MAML enables the autonomous systems to quickly learn and adapt to new environments or tasks with minimal data. In healthcare, DGS-MAML can help models better adapt to new patient data, facilitating personalized treatment recommendations and in biomedical sciences \cite{woessner2024identifying}. It is also useful for few-shot learning,  and improved generalization techniques to analyze rare diseases and experimental data, especially when experiments are too complex or expensive to repeat. In finance, DGS-MAML allows models to swiftly adjust to new market conditions or products, improving decision-making with limited historical data. Additionally, DGS-MAML can be applied to enhance the explainability of models \cite{anjum2024novel} by helping in transfering of knowledge and information between more complex and more accurate models to lower complexity less accurate models but with higher explainability and trust \cite{kinateder2025novel, matsuyama2025adaptive}.

\section{Conclusion}

In this work, we proposed DGS-MAML, a novel bi-level optimization algorithm that integrates gradient matching and domain generalization into the MAML framework. We extended the convergence analysis of SAGM—previously studied only in single-level settings—to the bi-level context. Our analysis demonstrated that DGS-MAML achieves a better convergence rate with respect to $T$ and offers a better PAC-Bayes generalization bound than SharpMAML.

Empirical evaluations on benchmark datasets confirmed that combining gradient matching with domain generalization leads to improved accuracy and better generalization, particularly in data-scarce scenarios. Applications range from natural language processing and cyberbullying detection to autonomous systems, healthcare, and finance, where DGS-MAML enables quick adaptation without full retraining.

In future work, we plan to explore the integration of other SAM-like algorithms within the MAML framework to further enhance performance and generalization. We would also like to explore other data domains and apply DGS-MAML to other models like regression and reinforcement learning.

\bibliography{maml_references}

\newpage
\onecolumn
\section{Appendix}

\subsection{Lemmas} 

In this section, we show the Lemmas necessary for the formulation of our proofs.

Let $k = C\sqrt{d}(1 + \frac{\alpha}{C\sqrt{d}} - \delta)$ in the following.
\newline
\newline
\textit{Lemma 1: Bounding $\mathbb{E}[\langle \widetilde{\nabla} \mathcal{L}(\theta_{t}), \widetilde{\nabla} \mathcal{L}_{SAGM} \rangle | \mathcal{F}_{t}]$}.
\begin{align}
    & \mathbb{E}[\langle \widetilde{\nabla} \mathcal{L}(\theta_{t}), \widetilde{\nabla} \mathcal{L}_{SAGM} \rangle | \mathcal{F}_{t}] \leq \mathbb{E}[ \langle \widetilde{\nabla} \mathcal{L}(\theta_t), \widetilde{\nabla} \mathcal{L}(\theta_{t}) \rangle \nonumber \\ 
    & + \langle \widetilde{\nabla} \mathcal{L}(\theta_t), \widetilde{\nabla} \mathcal{L}_{p}(\theta_t) \rangle \, | \, 
    \mathcal{F}_{t}] \\ 
    & \leq \mathbb{E}[||\widetilde{\nabla} \mathcal{L}(\theta_{t})||^2 \, | \, \mathcal{F}_{t}] + \mathbb{E}[||\widetilde{\nabla} \mathcal{L}|| \cdot ||\widetilde{\nabla} \mathcal{L}_{p}|| \, | \, \mathcal{F}_{t}] \\ 
    & = || \widetilde{\nabla} \mathcal{L} ||^2 + ||\widetilde{\nabla} \mathcal{L}(\theta_{t})|| \cdot ||\widetilde{\nabla} \mathcal{L}_{p}(\theta_{t})|| \,  \\
    & \leq C^2d + \sigma_{1}^2 + \sqrt{(C^2d + \sigma_{1}^2)(k^2 + \sigma_{2}^2)}
\end{align}
This completes the proof.

\textit{Lemma 2: Bounding $\frac{L\gamma^2}{2}\mathbb{E}[|| \widetilde{\nabla}\mathcal{L}_{SAGM}||^2 \, | \, \mathcal{F}_{t}]$}.

\begin{align}
    & \frac{L\gamma^2}{2}\mathbb{E}[|| \widetilde{\nabla}\mathcal{L}_{SAGM}||^2 \, | \, \mathcal{F}_{t}]  = \frac{L \gamma^2}{2} \mathbb{E}[||\widetilde{\nabla} \mathcal{L}(\theta_{t}) + \widetilde{\nabla}\mathcal{L}_{p}(\theta_{t})||^2 \, | \, \mathcal{F}_{t}] \\
    & = L \gamma^2\mathbb{E}[2||\widetilde{\nabla} \mathcal{L}(\theta_{t})||^2 + 2||\widetilde{\nabla}\mathcal{L}_{p}||^2 - ||\widetilde{\nabla} \mathcal{L}_{p} - \widetilde{\nabla}\mathcal{L}(\theta_{t})||^2 \, | \, \mathcal{F}_{t}] \\
    & = L \gamma^2(2 \mathbb{E}||\nabla\mathcal{L}||^2 + 2\mathbb{E}||\widetilde{\nabla} \mathcal{L}_{p}||^2 - \mathbb{E}[||\widetilde{\nabla}\mathcal{L}_{p} - \widetilde{\nabla} \mathcal{L}(\theta_{t})||^2 \, | \, \mathcal{F}_{t}]) \\
    & \leq L \gamma^2(2(C^2d + \sigma_{1}^2) + 2(k^2 + \sigma_{2}^2) - \mathbb{E}[||\widetilde{\nabla}h||^2 \, | \, \mathcal{F}_{t}]
\end{align}

By Lemma 5, we now have:
\begin{align}
    & \leq L \gamma^2(2(C^2d + \sigma_{1}^2) + 2(k^2 + \sigma_{2}^2) \nonumber \\ 
    & - (C^2d + k^2 + \sigma_1^2 + \sigma_2^2 - 2\sqrt{(C^2d + \sigma_1^2)(k^2 + \sigma_2^2)})   \\
    & = L \gamma^2(C^2 d + \sigma_{1}^2 + \sigma_{2}^2 + k^2 + 2\sqrt{(C^2d + \sigma_1^2)(k^2 + \sigma_2^2)})
\end{align}
This completes the proof.

\textit{Lemma 3: Bounding $||\nabla \mathcal{L}_{p}||$}. \\
Since $||\nabla\mathcal{L}|| \leq C\sqrt{d}$ by assumption 2, then:

\begin{align}
    & ||\nabla \mathcal{L}_{p}(\theta_{t})|| = ||\nabla \mathcal{L}(\theta_{t} + \hat{\varepsilon} - \delta \nabla \mathcal{L}(\theta_{t})) || \\
    & \leq \sqrt{(C + \frac{\alpha C}{C \sqrt{d}} - \delta C)^2 + ...} \\ 
    & \leq \sqrt{(C + \frac{\alpha}{\sqrt{d}} - \delta C)^2 d} \\ 
    & \leq C \sqrt{d} (1 + \frac{\alpha}{C\sqrt{d}} - \delta)
\end{align}
So $||\nabla \mathcal{L}_{p}|| \leq k$, which completes the proof.

\textit{Lemma 4: Bounding $||\nabla h||^2$:}
\begin{align}
    & ||\nabla h|| = ||\nabla \mathcal{L}_{p}(\theta) -\nabla \mathcal{L}(\theta)|| \\
    & ||\nabla h||^2 = ||\nabla \mathcal{L}_{p}(\theta) - \nabla \mathcal{L}(\theta)||^2 \\
    & = 2||\nabla \mathcal{L}{p}||^2 + 2||\nabla\mathcal{L}||^2 - ||\nabla \mathcal{L}{p} + \nabla \mathcal{L}||^2 \\
    & \leq 2C\sqrt{d}(1 + \frac{\alpha}{C\sqrt{d}} - \delta) \nonumber \\
    & + 2C^2d - (||\nabla \mathcal{L}{p}||^2 + 2||\nabla \mathcal{L}{p}|| \cdot || \nabla \mathcal{L}|| + ||\nabla \mathcal{L}||^2) \\ 
    & \leq 2k^2  + 2C^2d - k^2 - 2kC\sqrt{d} - C^2d \\ 
    & = k^2 - 2kC\sqrt{d} + C^2d
\end{align}
This completes the proof.

\textit{Lemma 5: \\
$||\widetilde{\nabla} h||^2 \leq C^2d + k^2 + \sigma_{1}^2 + \sigma_{2}^2 - 2\sqrt{(C^2d + \sigma_{1}^2)(k^2 + \sigma_{2}^2)}$}
\begin{align}
    & ||\widetilde{\nabla}h||^2 = ||\widetilde{\nabla}\mathcal{L}_{p} - \widetilde{\nabla}\mathcal{L}_{p}||^2 \\
    & = 2||\widetilde{\nabla}\mathcal{L}_{p}||^2  + 2||\widetilde{\nabla}\mathcal{L}||^2 - ||\widetilde{\nabla}\mathcal{L}_{p} - \widetilde{\nabla}\mathcal{L}||^2 \\ 
    & \leq 2(C^2d + \sigma_{1}^2) + 2(k^2 + \sigma_{2}^2) - ||\widetilde{\nabla}\mathcal{L}_{p} + \widetilde{\nabla}\mathcal{L}||^2 \\
    & \leq 2(C^2d + k^2) + 2(\sigma_{1}^2 + \sigma_{2}^2) \nonumber \\
    & - (||\widetilde{\nabla}\mathcal{L}_{p}||^2 + 2||\widetilde{\nabla}\mathcal{L}_{p}|| \cdot ||\widetilde{\nabla}\mathcal{L}|| + ||\widetilde{\nabla}\mathcal{L}||^2) \\
    & \leq C^2d + k^2 + \sigma_{1}^2 + \sigma_{2}^2 - 2\sqrt{(C^2d + \sigma_{1}^2)(k^2 + \sigma_{2}^2)}
\end{align}

This completes the proof.

\end{document}